\documentclass[letterpaper]{article}
\usepackage{}\def\year{2022}\relax
\usepackage{aaai22}
\usepackage{times}
\usepackage{helvet}
\usepackage{courier,dsfont}
\usepackage[hyphens]{url}
\usepackage{graphicx,color}
\usepackage{natbib}
\usepackage{caption}
\DeclareCaptionStyle{ruled}{labelfont=normalfont,labelsep=colon,strut=off}
\usepackage{algorithm}

\usepackage{algpseudocode}

\usepackage{lineno}

\usepackage{amsmath}
\usepackage{amsfonts}
\usepackage{subfigure}

\DeclareMathOperator*{\argmin}{arg\,min}

\usepackage{newfloat}

\usepackage{listings}
\floatstyle{ruled}
\newfloat{listing}{tb}{lst}{}
\floatname{listing}{Listing}

\title{Learning to Liquidate Forex: Optimal Stopping via Adaptive Top-$K$ Regression}
\author{Diksha Garg, Pankaj Malhotra, Anil Bhatia, Sanjay Bhat,  Lovekesh Vig, Gautam Shroff}
\affiliations{
(diksha.7, malhotra.pankaj, anil.bhatia, sanjay.bhat, lovekesh.vig, gautam.shroff)@tcs.com\\
TCS Research, New Delhi, India
}

\usepackage{bibentry}
\begin{document}

\maketitle

\begin{abstract}
We consider learning a trading agent acting on behalf of the treasury of a firm earning revenue in a foreign currency (FC) and incurring expenses in the home currency (HC). The goal of the agent is to maximize the expected HC at the end of the trading episode (e.g. over a financial quarter) by deciding to hold or sell the FC at each time step in the trading episode. We pose this as an optimization problem, and consider a broad spectrum of approaches with the learning component ranging from supervised to imitation to reinforcement learning. We observe that most of the approaches considered struggle to improve upon simple heuristic baselines. We identify two key aspects of the problem that render standard solutions ineffective - {\em i}) while good forecasts of future FX rates can be highly effective in guiding good decisions, forecasting FX rates is difficult, and erroneous estimates tend to degrade the performance of trading agents instead of improving it, {\em ii}) the inherent non-stationary nature of FX rates renders a fixed decision-threshold highly ineffective.
To address these problems, we propose a novel supervised learning approach that learns to forecast the top-$K$ future FX rates instead of forecasting all the future FX rates, and bases the hold-versus-sell decision on the forecasts (e.g. hold if future FX rate is higher than current FX rate, sell otherwise).
Furthermore, to handle the non-stationarity in the FX rates data which poses challenges to the i.i.d. assumption in supervised learning methods, we propose to adaptively learn decision-thresholds based on recent historical episodes.
Through extensive empirical evaluation across 21 approaches and 7 currency pairs, we show that our approach is the only approach which is able to consistently improve upon a simple heuristic baseline. Further experiments show the inefficacy of state-of-the-art statistical and deep-learning-based forecasting methods as they degrade the performance of the trading agent.
\end{abstract}

\section{Introduction}
We consider learning a trading agent acting on behalf of the treasury of a firm. The firm earns revenue in a foreign currency (FC) but incurs expenses in the home currency (HC). This requires converting FC to HC in a currency exchange (FX) where the rates for exchanging the currencies, i.e. FX rates, vary over time. The goal of the trading agent is to decide to \textit{sell} or \textit{hold} the FC at each timestep so as to maximize the HC balance at the end of an episode.

Consider an oracle agent that has access to day-level future FX rates. Then, the decision to hold or sell is simple: ``hold'' if a future FX rate is higher than today's rate, and ``sell'' if all future FX rates are lower than today's. 
Learning-based agents have been successfully developed for forex trading, e.g. based on support vector machines, random forests \cite{dunis2002modelling}, neural networks \cite{dunis2011higher}, reinforcement learning (RL) \cite{dempster2006automated}, etc.
Most of these methods first learn to forecast the FX rates, and then use them for decision-making.
Given the complexity and inherent stochasticity of FX rate time-series, forecasting intraday or day-level FX rates has been known to be a notoriously hard problem \cite{meese1983empirical,austin2004adaptive}.

In this work, we argue that a more natural objective for the learning task could be to forecast the maximum or top-$K$ future FX rates, and to use those for decision-making instead of relying on the forecasts for all/consecutive time steps over a horizon.
We motivate this in context of an optimal stopping formulation \cite{longstaff2001valuing} of the problem while also indicating shortcomings of other potential learning objectives such as those based on deep RL, imitation learning (IL), classification, etc.
\textbf{The key contributions of this work are:}
\textbf{1.} We provide a formulation of the FC selling problem as an optimal stopping problem. 
\textbf{2.} We benchmark several traditional as well as learning-based approaches for solving this problem: i. basic forex indicators traditionally used by traders, ii. supervised learning methods based on classification (sell-versus-hold) and regression
formulations, iii. several variants of RL and IL, and iv. dynamic programming (DP) with function approximation. We discuss the limitations of these approaches in the context of optimal stopping problems, and empirically prove the inefficacy of these methods.
\textbf{3.} We rely on the following novel insights to overcome the highlighted limitations of existing methods: i. \textbf{estimating the maximum (or top-$K$) future FX rate(s)} is better for decision-making instead of forecasting several future FX rates. 
Additionally, estimating the maximum (or top-$K$) future FX rate(s) is closely related to the actual decision-making problem.
On the other hand, using 
forecasted FX rates for subsequent decision-making would typically require further non-trivial processing by the agent. 
ii. \textbf{adaptive local decision-thresholds} using recent historical FX trends are better than using a global decision-threshold obtained using the entire training set. 
We show that our approach using top-$K$ FX rate estimates and adaptive local decision-thresholds significantly improves upon 20 traditional and learning-based baselines considered.
    
\section{Related Work}
A large body of literature suggests using multi-step forecasting of future FX rates for subsequent forex trading decisions. However, due to inherent stochasticity and complexity of FX rates market, forecasting FX rates has been known to be a challenging and tedious problem \cite{meese1983empirical,austin2004adaptive,shah2021forecasting, duan2021learning}. 
In this work, we consider an alternative to such standard forecasting approaches where we learn to forecast the top-$K$ future FX rates for subsequent trading decisions. Through extensive empirical evaluation, we show the benefits of forecasting top-$K$ future FX rates over standard multi-step forecasting of future FX rates. 

Furthermore, we show that our approach based on top-$K$ FX forecasts can be seen as an approximate solution to the forex trading problem when mapped to an Optimal Stopping formulation \cite{longstaff2001valuing}. Optimal Stopping-based solutions have been successfully considered for several problems prevalent in the finance domain including options pricing \cite{becker2019deep}, stock-market price predictions \cite{griffeath1974optimal}, and portfolio optimization \cite{choi2004optimal}. 
However, to the best of our knowledge, Optimal Stopping-based solution for forex trading has not been proposed yet. In this work, we take a step towards bridging this gap by mapping the FX trading problem to Optimal Stopping, and discuss and evaluate solutions based on several existing formulations and the proposed formulation. Among the existing formulations, we consider implementations based on Q-learning \cite{herrera2021optimal, fathan2021deep}, backward recursion \cite{herrera2021optimal,kohler2010pricing}, and DP with function approximation \cite{yu2007q}.
 We show that the implementations of existing formulations struggle to improve upon a simple heuristic-based baseline we propose, and then provide a novel solution based on top-$K$ FX forecasts that improves upon all baselines considered.

\section{Foreign Currency Liquidation  Problem \label{sec:trading}}

Consider a trading agent  acting on behalf of the treasury of a firm earning revenue in FC and incurring expenses in HC. The firm receives random FC revenues $R_{0},\ldots,R_{T}$ at time periods $0,\ldots,T$, respectively. The treasury needs to convert the FC to HC in a FX market where the exchange rate at time $t=0,\ldots,T$ is $X_{t}$ units of HC per unit of FC.  The agent's task is to determine quantities $\Delta_{0},\ldots,\Delta_{T}$ of FC to exchange over the horizon $0,\ldots,T$ such that all the FC received over the horizon is exchanged, and the expected HC earned through the exercise is a maximum. The agent's decisions are subject to the obvious practically motivated constraint that the decision $\Delta_{t}$ at time $t$ should not depend on the future FX rate $X_{s}$ for any $s>t$.  We write this as  $\Delta_{t} \sim \mathcal{F}_{t}$ for every $t$, where  $\mathcal{F}_{t}$ represents the information available for decision-making up to time $t$. We additionally assume a no-short-selling constraint requiring that the cumulative FC sold in the FX market up to any time does not exceed the cumulative FC earned up to that time. 
More formally, the agent's optimization problem is given by
$
\max_{\{\Delta_{t}\}_{t=0}^{T}}\mathbb{E}\left(\sum_{t=0}^{T}\Delta_{t}X_{t}\right) \label{objeq}
$,
subject to the constraints:  $\Delta_{t} \sim \mathcal{F}_{t}$  and $\sum_{l=0}^{t}\Delta_{l} \leq \sum_{l=0}^{t}R_{l}$ for each $t=0,\ldots,T$, and $\sum_{l=0}^{T}\Delta_{l}=  \sum_{l=0}^{T}R_{l}$. This optimization problem appears challenging because of the presence of two stochastic processes and a combination of  informational and pointwise constraints. 
However, the commonplace trader's dictum ``buy low, sell high" suggests that the agent should hold all FC received till the time when the the FX rate reaches a maximum, sell all accumulated FC at that time, and then repeat the same strategy till the end of the horizon. In other words, the agent's problem is one of timing the market. This insight forms the basis for alternative formulations and solution approaches, which we  present next. 
\section{Overview\label{sec:AT}}
In the next section, we will present a few known approaches as well as our proposed approach for solving the FX trading agent's problem of ``timing the market''. In essence, all the approaches that we present involve estimating, at each time $t$, a decision variable $d_{t}$ representing the difference between the best possible future FX rate and the current rate, and comparing it with a decision boundary in the form of an adaptive threshold. 

\textbf{Estimating the Decision Variable:} 
In the first approach, we cast the problem as a general, non-Markovian optimal stopping problem, and apply recursive regression in the spirit of \citet{longstaff2001valuing} to estimate $d_{t}$. Unfortunately, the recursive regression requires as many regression models as the number of time steps. Hence we next use  dynamic-programming-based value-function approximation in a finite-horizon  Markovian framework to estimate $d_{t}$. We also consider value-function approximation and $Q$-learning for optimal stopping \cite{bertsekas1995neuro} using a Deep Q-Network-type \cite{van2016deep} algorithm in a infinite-horizon, stationary Markovian setting to obtain estimators for  $d_{t}$ that do not explicitly depend on time. Our proposed approach is designed to mitigate the problem of overestimation bias inherent in the previous approaches, and involves using estimates of top-$K$ future FX rates to build an estimate of $d_{t}$.  

\textbf{Adaptive Threshold (AT):} Most traditional as well as learning-based agents apply a threshold over an estimate for final sell/hold decision for handling bias in  estimates. Hence, instead of comparing $d_{t}$ with $0$ to trigger a sell decision, we compare it with a threshold $\delta^{\mathrm{e}}$ computed at the start of each episode $\mathrm{e}$. The condition $d_{t}-\delta^{\mathrm{e}}<0$ then triggers the decision to sell all available FC at time $t$ in the episode $\mathrm{e}$. 
In its simplest form, the threshold $\delta^{\mathrm{e}}$ can be chosen to be a constant independent of $\mathrm{e}$. However, given the non-stationary nature of the FX rates data, we use an episode-dependent threshold chosen from a finite set of candidates. Each candidate threshold is used along with the chosen estimator for $d_{t}$ for decision-making on a collection of historical episodes immediately preceding the episode $\mathrm{e}$. The candidate threshold that yields the highest average cumulative payoff on the historical episodes is chosen as the AT for the episode $\mathrm{e}$.

\section{Proposed Solution \label{sec:formulation}}
We describe different formulations and instantiate solutions to the above-mentioned generic problem. We highlight practical challenges in the solutions to these formulations and motivate our proposed solution.
Without loss of generality,  unless stated otherwise, all solutions use deep feedforward neural networks as the function approximators. Also, in all instances, we consider selling all FC available till time $t$ in case of ``sell'' decision at time $t$.
F$i$ and I$i$ refer to $i$th formulation and corresponding implementation, respectively.

\subsection{F1: Optimal Stopping Formulation}
The ``sell high"  strategy outlined above cannot be implemented in practice as it requires knowing the future evolution of the FX rate. Nevertheless, it indicates that the optimal solution must be in terms of decision rules which use the information available up to time $t$ to recommend if one should sell the FC at $t$ or continue to hold it. In other words, the problem defined in Section \ref{sec:trading}
may be solved if, for each $t=0,\ldots,T$, we can find a stopping time $\tau_{t}\geq t$ such that $X_{\tau_{t}}=\max_{\sigma \geq t} \mathbb{E}[X_{\sigma}|\mathcal{F}_{t}]$, 
where the maximum is pointwise over all stopping times $\sigma \geq t$. The optimal decision rule for time $t$  then is to sell all accumulated FC at $t$ if and only if the condition $t=\tau_{t}$ is satisfied. The stopping time $\tau_{t}$ can be obtained through the solution of the  standard optimal stopping problem of finding $\max\mathbb{E}(X_{\sigma})$, where the maximum is over all stopping times $\sigma$ in the range $0,\ldots,T$. The solution is in terms of the {\em Snell envelope} $\{Y_{t}\}_{t=0}^{T}$ of the FX rate process defined through the backward recursion 
$Y_{T}=X_{T}$ and $Y_{t}=\max\{X_{t},\mathbb{E}(Y_{t+1}|\mathcal{F}_{t})\}$ for each $t=1,\ldots,T$.
The optimal decision rule at time $t$ is then to sell all FC in hand if and only if $X_{t} \geq Y_{t}$.

\subsubsection{I1: Backward Recursion using Regression}
The backward recursion
given in F1 cannot be solved analytically unless statistics of the FX rate process are known in analytical form, and data-driven techniques are required for finding approximate solutions to $Y_T$ and $Y_t$. 
 We adapt the well-known approach by \citet{longstaff2001valuing} for obtaining approximations to $Y_T$ and $Y_t$ using data. This is achieved by training one neural network for each $t=1,\ldots,T$ such that the neural network for time $t$ yields an approximation $\hat{Y}_{t}(\mathbf{f}_{t-1},\theta_{t})$ to the conditional expectation $\mathbb{E}(Y_{t}|\mathcal{F}_{t-1})$, where $\mathbf{f}_{t-1}$ represents the feature vector at time $t-1$, and $\theta_{t}$ represents the neural network parameters. Assuming that the neural network for time $t+1$ is trained, the neural network for time $t=0,\ldots,T-1$ is trained by finding $\theta_{t}=\argmin_{\theta} \mathcal{L}_{t}(\theta)$ for  the loss function $\mathcal{L}_{t}(\theta)=\sum[\hat{Y}_{t}(\mathbf{f}_{t-1},\theta)-\max\{X_{t},\hat{Y}_{t+1}(\mathbf{f}_{t},\theta_{t+1})\}]^{2}$, where the sum is over a set of training samples for the triplet  $(\mathbf{f}_{t-1},\mathbf{f}_{t},X_{t})$. Since $Y_T=X_T$, the target in the loss  function $\mathcal{L}_{T}$ is $X_{T}$ itself. 
Given neural networks trained as above, the recommended action at time $t$ given the feature vector $\mathbf{f}_{t}$ is decided using $d_t= [\hat{Y}_{t+1}(\mathbf{f}_{t},\theta_{t+1}) - X_{t}]$.

Since this approach involves training a different neural network for each time step, the training process becomes computationally  intensive as the time horizon or the number of features increases. This problem can be mitigated to a certain extent by invoking additional assumptions.

\subsection{F2.1: Finite-Horizon Markov Chain Formulation}
Our proposed approach for approximately solving the backward recursion in F1 uses the Markovian framework. More precisely, we assume that there exists a Markov chain (MC) $\{S_{t}\}_{t=0}^{T}$ evolving on a state space $\mathcal{S}$, and an observation function $h:\mathcal{S}\rightarrow (0,\infty)$ such that $X_{t}=h(S_{t})$ for all $t$. In this case, the Snell envelope described in F1 
takes the form  $Y_{t}=V_{t}(S_{t})$ for each $t=0,\ldots,T$, where the sequence of {\em value functions} $V_{0}, \ldots,V_{T}$ on $\mathcal{S}$ is defined  by the finite-horizon DP equation $V_{T}(s)=h(s)$ and $V_{t}(s)= \max\{h(s),\mathbb{E}[V_{t+1}(S_{t+1})|S_{t}=s]\}$ 
for all $s\in\mathcal{S}$. 
The optimal decision whenever in state $s$ at time $t$  is to sell all FC in hand if and only if $h(s) \geq V_{t}(s)$.

\subsubsection{I2.1: Dynamic Programming (DP) with Function Approximation} 
To apply the approach described above, the state is chosen to be the collection of features that are expected to be relevant for the decision-making (see Section 6). We train a neural network such that its output  $\hat{V}(t,s,\theta)$ at time $t$ and state $s$ approximates $\mathbb{E}[V_{t+1}(S_{t+1})|S_{t}=s]$, where $\theta$ represents the neural network parameters.
 The neural network is trained by minimizing the loss function $\mathcal{L}(\theta)=\sum[\hat{V}(t,s,\theta)-\max\{h(s^{\prime}),\hat{V}(t+1,s^{\prime},\theta^{\prime})\}]^{2}$, where the sum is over samples of current state $s$, current time $t$, and next state $s^{\prime}$, and $\theta^{\prime}$ are the parameters of a target neural network as used in the DQN-algorithm \cite{van2016deep}. Note that for terminal samples, that is, for $t=T$, the estimate $\hat{V}(t+1,s^{\prime},\theta^{\prime})$ in the loss function is set to $0$ to be consistent with the dynamic program in F2.1. 
On the completion of training, the recommended decision when in state $s$ at time $t$ is chosen using $d_t=[\hat{V}(t,s,\theta) - h(s)]$.
Note that the function $h$ mapping features to the current FX rate is known in our case, and hence not approximated.

\subsection{F2.2: Infinite Horizon MC Formulation}
In cases where $T$ is large, we may further approximate the solution of the dynamic program in F2.1 by considering the limit $T\rightarrow \infty$ and assuming that the MC $\{S_{t}\}_{t=0}^{\infty}$ is stationary. In this case, the value $V(s)$ of taking the optimal decision at a state $s\in\mathcal{S}$ is independent of time, and the  value function 
satisfies the Bellman equation $V(s)=\max\{h(s),\mathbb{E}[V(S_{t+1})|S_{t}=s]\}$.
The optimal decision whenever in state $s$ is to sell all FC in hand at if and only if $h(s) \geq V(s)$. 

\subsubsection{I2.2: Value Function Approximation}
As in subsection I2.1, we  use a neural network to obtain an approximation $\hat{V}(s,\theta)$ to $\mathbb{E}[V(S_{t+1})|S_{t}=s]$, where $\theta$ represents the neural network parameters. We train the neural network using a DQN-like algorithm using the loss function given by
$\mathcal{L}(\theta)=\sum [ \hat{V}(s,\theta) - \max \{ h(s^{\prime}),\hat{V}(s^{\prime},\theta^{\prime}) \}]^2$, where the sum is over samples of transitions $s\mapsto s^{\prime}$, and  $\theta^{\prime}$ represents the parameters of a target network updated as in the DQN algorithm. 
Upon the completion of training, the recommended decision whenever in state $s$ is decided using $d_t=[\hat{V}(s,\theta) - h(s)]$.

\subsection{F2.3: $Q$-function Formulation}
The Bellman equation in F2.2 can be rewritten using the $Q$ function notation. To this end, we introduce a binary-valued action $a$ such that $a=1$ represents the decision to sell all FC in hand, while $a=0$ represents the decision to continue holding FC.  The optimal $Q$ function $Q:\mathcal{S}\times\{0,1\}\rightarrow (0,\infty)$ is defined such that $Q(s,a)$ is the value of taking action $a$ in state $s$ followed by optimal actions thereafter. Thus, $Q(s,a)$ equals $h(s)$ if $a=1$, and equals $\mathbb{E}[V(S_{t+1})|S_{t}=s]$ if $a=0$. It is then clear that $V(s)=\max_{a\in\{0,1\}}Q(s,a)$ for all $s\in\mathcal{S}$. Consequently, the Bellman equation for the optimal $Q$ function is given by $Q(s,a)=h(s)$ for $a=1$, and $Q(s,a)=\mathbb{E}[\left.\max_{a^{\prime}\in\{0,1\}}Q(S_{t+1},a^{\prime})\right|S_{t}=s]$ for $a=0$,
for all $s\in\mathcal{S}$. The optimal action at a state $s$ is then the action which yields a maximum for $Q(s,\cdot)$.

\subsubsection{I2.3: $Q$-Learning for Optimal Stopping}
To obtain an approximation for the $Q$ function in F2.3, 
we use a neural network to obtain an approximation $\hat{Q}(s,a,\theta)$ to $Q(s,a)$, where $\theta$ represents the neural network parameters. The neural network is trained in a similar fashion as the DQN-algorithm using the loss function 
$\mathcal{L}(\theta)=\sum[\hat{Q}(s,a,\theta)-\mathds{1}_{\{a=1\}}h(s)-\mathds{1}_{\{a=0\}}\max_{a^{\prime}}\hat{Q}(s^{\prime},a^{\prime},\theta^{\prime})]^{2},$ where the sum is over sample transitions $(s,a)\mapsto s^{\prime}$, and  $\theta^{\prime}$ are the parameters of a target network as before. Once trained, the recommended action at a state $s$ is decided using $d_t = [\hat{Q}(s,0,\theta) - h(s)]$.
The function approximations I2.1-2.3 attempt to  estimate  the conditional expectation of a maximum
of two quantities by averaging over samples of the maximum of those two quantities. For instance, I.2.2 approximates $\mathbb{E}[V(S_{t+1})|S_{t}=s]$ by using approximate samples for 
$\max\{h(S_{t+1}),V(S_{t+1})\}$.  It is well known that using this kind of a sample-based estimate for a maximum over expectations typically leads  to overestimation bias \cite{fujimoto2019off, fujimoto2019benchmarking, pentaliotis2020investigating}.  
We next propose an approach that simultaneously estimates multiple future FX rate values in a multi-task learning setup, and overcomes the overestimation bias issue.

\subsection{F2.4: Multi-task Supervised Learning Formulation (Proposed)}
To describe the idea underlying the proposed approach, we introduce some notation. First, for each $t$ and each $i=1,\ldots,T-t$, let $X_{t+[i]}$ denote the $i$th largest FX rate with $X_{t+[1]}$ being the largest FX rate, of the collection $\{X_{t+1},\ldots,X_{T}\}$. Next, let $K<T$ be  a fixed integer,  
and define $Z_{t+1}=\frac{1}{K}(X_{t+[1]}+\cdots X_{t+[K]})$. Note that $Z_{t}$ is the average of the $K$ highest values of the FX rate from time $t+1$ onwards. Finally define
\begin{equation}
    W(s)=\max\{h(s),\mathbb{E}[Z_{t+1}|S_{t}=s]\}.   \label{topkeq}
\end{equation}
The recommended decision for state $s$ at time $t$ is to sell all accumulated FC if and only if $h(s) \geq W(s)$.
Our proposed approach is to estimate $W$ in (\ref{topkeq}) using data samples, and apply the recommended action according to the estimate. 

\subsubsection{I2.4: Proposed Solution: top-$K$ FX Rates Forecasting + Sell-AT (top-$K$ + AT)}
We consider a multi-task setting with $K$ tasks, where each task is a regression task corresponding to estimating one of the top-$K$ future FX rates. In other words, we consider a regression task where the target variable is $K$-dimensional. We use a neural network to obtain an approximation $\hat{W}(s,\theta)$ to $\mathbb{E}[Z_{t+1}|S_{t}=s]$, where $\theta$ represents the neural network parameters.
We consider the following weighted loss function based on the ranks of the future exchange rates for sample state $s$ at time step  $t$ in an episode:
\begin{equation}
 \mathcal{L}_{t}(\boldsymbol{\theta}) = \sum_{k=1}^K \frac{1}{k}   \left ( \hat{W}_k(s,\theta) -  X_{t+[k]} \right )^2.
 \label{topkloss}
\end{equation}
In practice, for $t>T-K$ the summation in (\ref{topkloss})  is over $T-t$ terms only. Finally, $\hat{W}(s,\theta) = \frac{1}{K}\sum_{k=1}^K\hat{W}_k(s,\theta)$.

Note that in contrast to I2.1-2.3, the target in the loss function is not an estimated value but rather obtained from direct observations. Hence, the agents learned using this objective are easier to train in practice as the targets are less noisy and the learning does not suffer from the overestimation bias.
Upon the completion of training, the recommended decision whenever in state $s$ is decided using $d_t=[\hat{W}(s,\theta) - h(s)]$.
Note that $K=1$ is a special case with single task with the maximum FX rate as a univariate target:
$
 \mathcal{L}_{t}(\boldsymbol{\theta}) = ( \hat{W}(s,\theta) -  \max \{ X_{t+1}\dots X_T   \}  )^2.
$

\section{Experimental Evaluation}

\begin{table}
    \centering
	\footnotesize
	\caption{Statistics of the datasets considered.  \label{tab:stats}}
	\scalebox{0.7}{
	\centering 
		\begin{tabular}{l|c|c|c}
		\hline
		Statistic &\multicolumn{1}{|c|}{\textbf{Train }} & \multicolumn{1}{|c|}{\textbf{Validation}} & \multicolumn{1}{|c}{\textbf{Test}} \\
		\hline
		\hline
		$\#$total episodes & 1001 & 250 & 250 \\
		$\#$unique episodes & 233 & 83 & 83 \\
		Start date (dd-mm-yy) & 03-01-11 & 10-01-17 & 25-04-19 \\
		End date (dd-mm-yy) & 05-01-17 & 22-04-19 & 31-08-21 \\
		\hline
	\end{tabular}}
\end{table}

\begin{figure}

	\subfigure[\scriptsize ]{\includegraphics[width=0.149\textwidth]{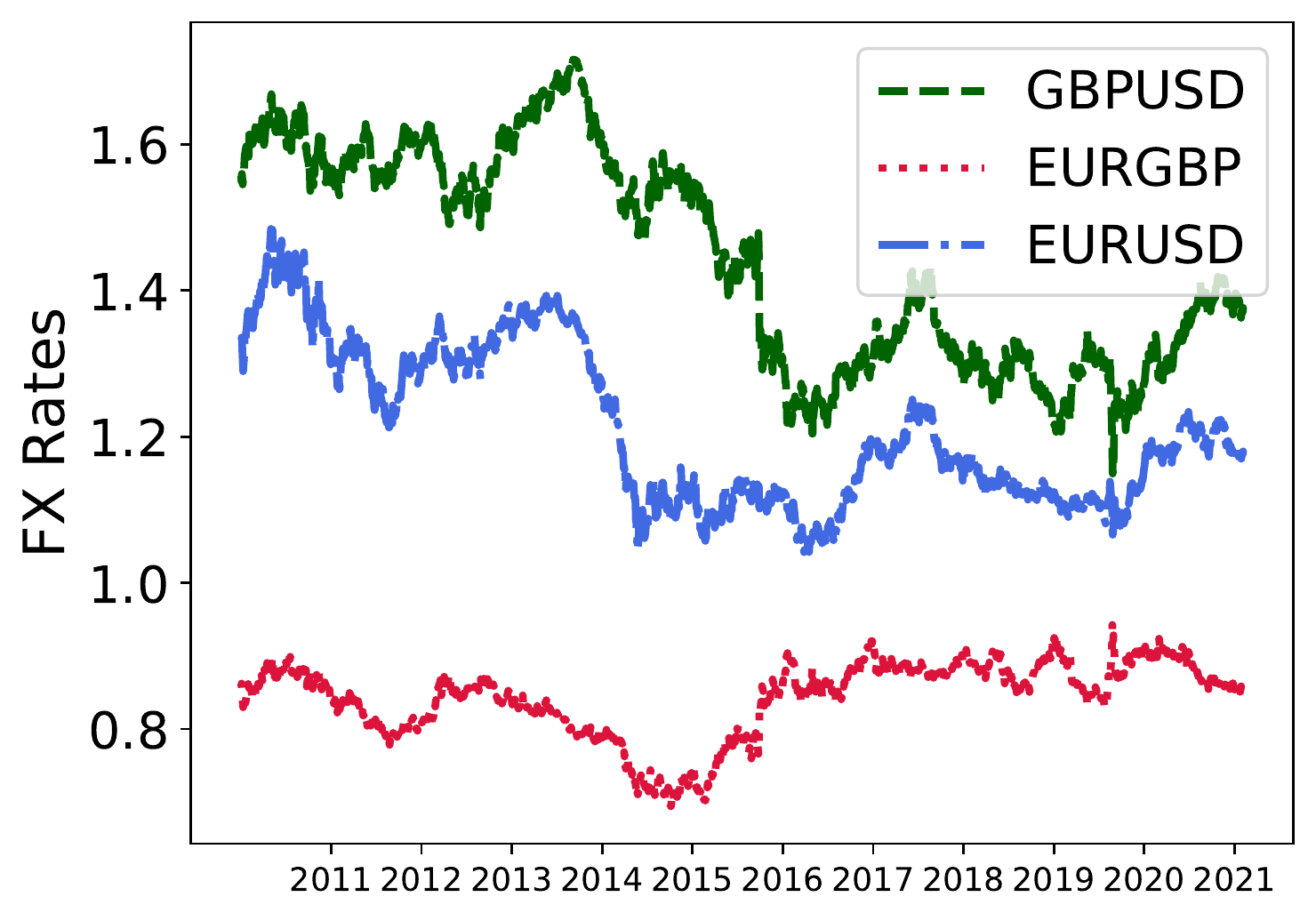}}
	\subfigure[\scriptsize ]{\includegraphics[width=0.15\textwidth]{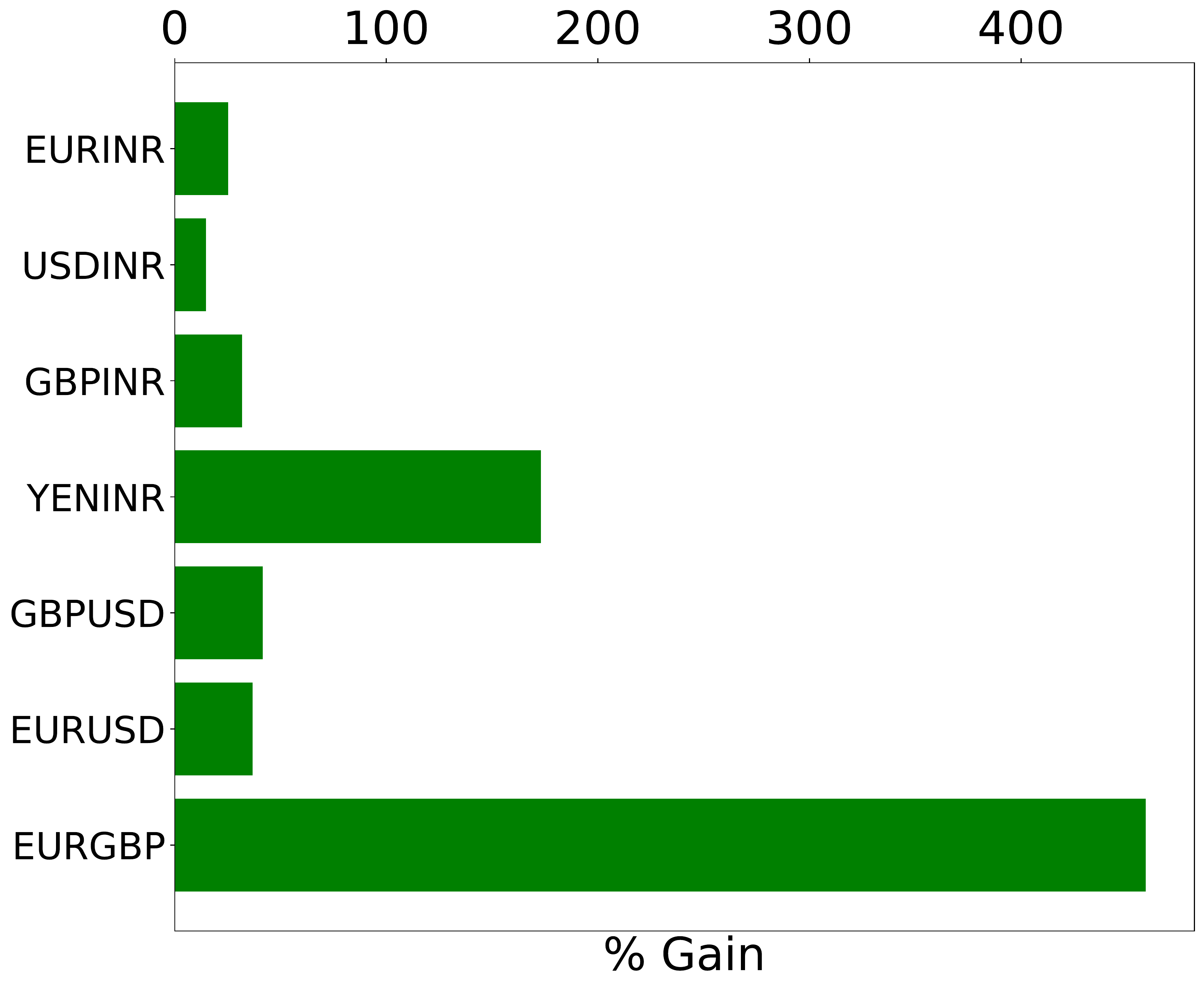}}
	\subfigure[\scriptsize ]{\includegraphics[width=0.137\textwidth]{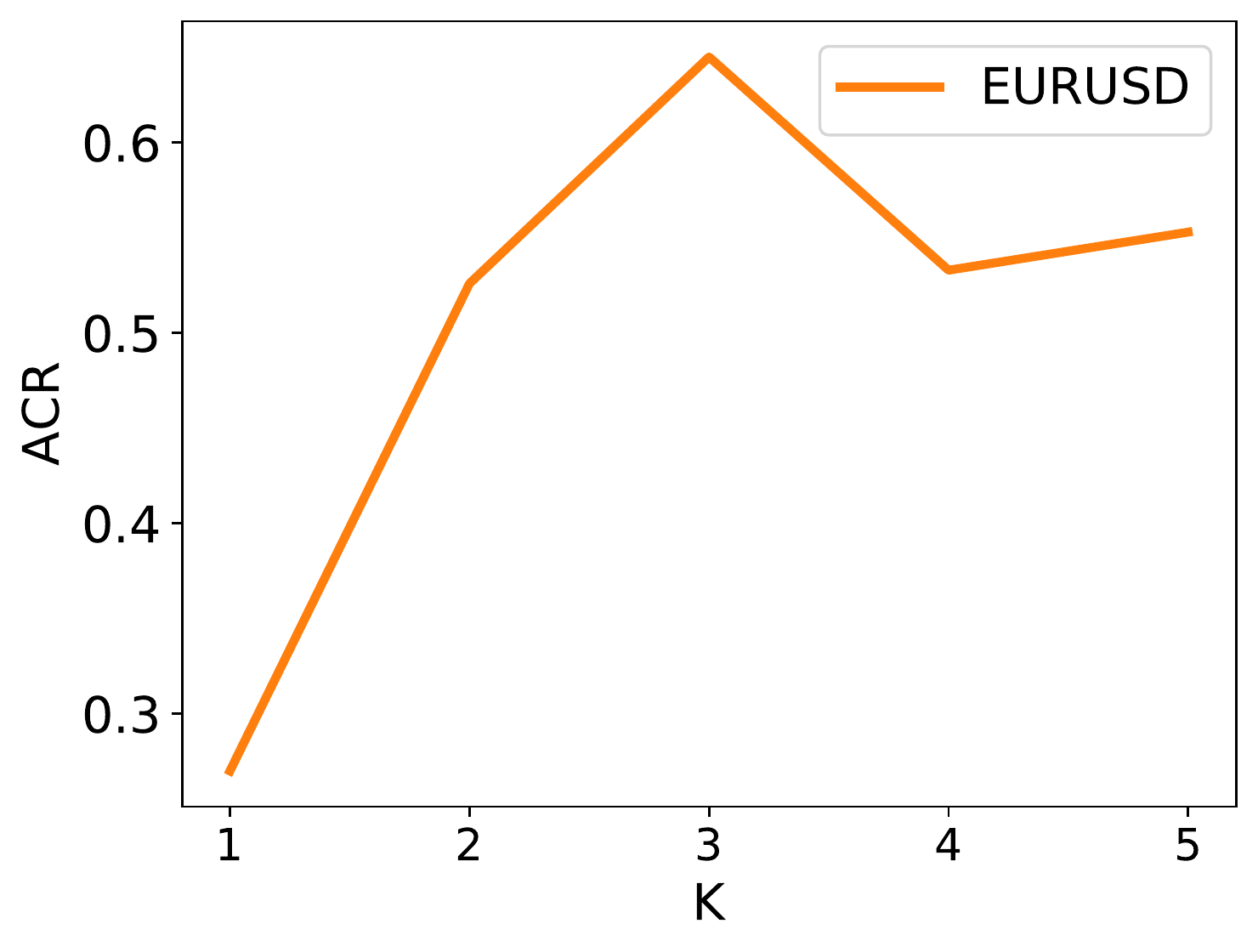}}
	
	\caption{\textbf{(a)} FX rates time-series of some of the currency pairs considered depicting inherent stochastic nature of FX data. \textbf{(b)} shows the \%age gain due to AT as a post-processing step instead of fixed threshold for our approach. \textbf{(c)} show the sensitivity of results to $K$ in our approach. For remaining currency pairs, we observe behaviour similar to EURUSD except for GBPINR.
	\label{fig:ranks_gains}}
\end{figure}

\begin{table*}
    \centering
	\footnotesize
	\caption{Performance comparison  of  various  approaches  with  the  proposed  top-$K$ + AT in terms of average cumulative rewards(ACR) and overall ranking across all currency pairs. \label{tab:performance}}
	
	\scalebox{0.67}{
	\centering 
		\begin{tabular}{l|c|c|c|c|c|c|c|c}
		\hline
		Method &\multicolumn{1}{|c|}{\textbf{EURGBP}} & \multicolumn{1}{|c|}{\textbf{EURUSD}} & \multicolumn{1}{|c|}{\textbf{GBPUSD}} & \multicolumn{1}{|c}{\textbf{YENINR}} & \multicolumn{1}{|c}{\textbf{GBPINR}} & \multicolumn{1}{|c}{\textbf{USDINR}} & \multicolumn{1}{|c}{\textbf{EURINR}} & \multicolumn{1}{|c}{\textbf{Rank}} \\
		\hline
		\hline
		\multicolumn{9}{c}{\textbf{Naive Baselines}}\\
		\hline
		Sell At End & -0.243 & 0.463 & 0.755 & 0.171 & 1.070 & 0.343 & 0.781 & 9.071 \\
		Sell Immediately & -0.070 & 0.270 & 0.370 & 0.143 & 0.530 & 0.180 & 0.441 & 19.143 \\
		Sell Greedily & -0.075 & 0.387 & 0.438 & 0.264 & 0.762 & 0.241 & 0.644 & 12.286 \\
		Sell-AT & 0.082 & \underline{0.643} & \underline{0.870} & \underline{0.461} & \underline{1.179} & \underline{0.476} & \textbf{0.999} & \underline{2.357} \\
	    \hline
	    \multicolumn{9}{c}{\textbf{Basic Forex Indicators +  Cross-over}}\\
	    \hline
	    EMA(10)$\And$EMA(20) & -0.227 & 0.374 & 0.418 & 0.095 & 0.715 & 0.250 & 0.638 & 15.500 \\
		Exchange rate$\And$EMA(10) & -0.170 & 0.385 & 0.515 & 0.139 & 0.636 & 0.235 & 0.559 & 15.071 \\
		Exchange rate$\And$EMA(100) & -0.198 & 0.514 & 0.755 & 0.150 & 1.067 & 0.343 & 0.781 & 8.643 \\
		EMA(50)$\And$EMA(100) & -0.227 & 0.446 & 0.619 & 0.231 & 1.073 & 0.319 & 0.514 & 10.643\\
		MACD$\And$signal & -0.053 & 0.312 & 0.474 & 0.228 & 0.690 & 0.201 & 0.580 & 13.286 \\
		MACD$\And$signal with MACD$>$0 & -0.034 & 0.397 & 0.420 & 0.407 & 0.748 & 0.391 & 0.751 & 8.857\\
		\hline
		\multicolumn{9}{c}{\textbf{F1: Dynamic Programming with Function Approximation (DP)}}\\
        \hline
		Backward Recursion w/ Regression (BRR)[I1] & -0.070 & 0.277 & 0.388 & 0.147 & 0.546 & 0.190 & 0.454 & 17.500 \\
        \hline
		\multicolumn{9}{c}{\textbf{Reinforcement Learning Using DQN}}\\
		\hline
		Vanilla & -0.031 & 0.366 & 0.458 & 0.293 & 0.664 & 0.208 & 0.559 & 12.643 \\
		+ ranking-based rewards  & -0.070 & 0.270 & 0.370 & 0.143 & 0.530 & 0.180 & 0.441 & 19.143 \\
		+ binary rewards & -0.023 & 0.456 & 0.388 & 0.167 & 0.575 & 0.216 & 0.649 & 12.786 \\
		\hline
        \multicolumn{9}{c}{\textbf{Imitation Learning (Classification Task)}}\\
		\hline
		Vanilla & -0.243 & 0.463 & 0.755 & 0.171 & 1.070 & 0.343 & 0.781 & 9.071 \\
		With downsampling & -0.050 & 0.509 & 0.442 & 0.191 & 1.070 & 0.443 & \underline{0.942} & 6.857 \\
		Using Focal Loss & -0.067 & 0.288 & 0.386 & 0.150 & 0.534 & 0.194 & 0.455 & 16.929 \\
		\hline
        
        \multicolumn{9}{c}{\textbf{F2.1-F2.3: Markov Chain Formulation}}\\
        \hline
		With Finite horizon [I2.1] & 0.107 & 0.435 & 0.522 & 0.356 & 0.952 & 0.360 & 0.730 & 7.143 \\
		With Infinite horizon [I2.2] & 0.102 & 0.492 & 0.552 & 0.349 & 0.883 & 0.341 & 0.798 & 6.571 \\
		Q-Learning for Optimal Stopping [I2.3]& \underline{0.112} & 0.492 & 0.552 & 0.349 & 0.883 & 0.374 & 0.729 & 6.429 \\
		
		\hline
		\multicolumn{9}{c}{\textbf{F2.4: Supervised Learning (Forecasting task)}}\\
		\hline
		\textbf{top-$K$ + AT (proposed) [I2.4]} & \textbf{0.318} & \textbf{0.645} & \textbf{1.105} & \textbf{0.475} & \textbf{1.248} & \textbf{0.573} & \textbf{0.999} & \textbf{1.071}\\ 
		\hline
		\hline
		Oracle & \textit{0.847} & \textit{1.194} & \textit{1.720} & \textit{1.245} & \textit{1.897} & \textit{0.942} & \textit{1.573} & --\\
		\hline
		
	\end{tabular}}
\end{table*}

We consider day-wise exchange rates for 7 currency pairs (combinations of INR, USD, EURO, and YEN), as shown in Fig. \ref{fig:ranks_gains}(a). The datasets for the 2010 to 2021 period are downloaded from Yahoo Finance. As shown in Table \ref{tab:stats}, we split the time-series of FX rates for each currency pair chronologically into three disjoint sets, i.e., train, validation and test. We use rolling windows with a shift of 5 days to create episodes of length $T=58$, which is one quarter of a year after removing weekends and holidays.

Consider the \textbf{Oracle Policy} of selling at time $t$ if $X_t \geq \max(X_{t+1},...X_{T})$, else holding. This policy  assumes access to all future FX rates in an episode, and thus provides an upper bound on the performance of any agent.
\textbf{Oracle-$n$} with $n < T$ is a sub-optimal oracle policy where only next $n$ FX rates at each time are known. Thus, \textbf{Oracle-$n$} sells at time $t$ if $X_t \geq \max(X_{t+1},...X_{t+n})$, else it holds.  
In addition to the approaches I2.1-I2.4, we consider following baselines for evaluation:
\subsubsection{Naive Baselines}
i. \textbf{Sell At End}:
Hold FC revenues $R_{0},\ldots,R_{T}$ till end of episode, i.e. time $T$, and sell all FC on day $T$.
ii. \textbf{Sell Immediately}: 
Sell FC revenue  $R_{t}$ on day $t$ itself. iii. \textbf{Sell Greedily}:
Sell at time $t$ if FX rate $X_t$ is greater than the first day FX rate $X_1$ of an episode, else hold. iv. 
\textbf{Sell if current FX rate $>$ Adaptive Threshold  (Sell-AT)}
Sell at time $t$ if $X_t > \delta^e$, where $\delta^e$ is the adaptive threshold obtained using same approach as mentioned before, with the only difference that decisions are made by directly comparing FX rate only with a threshold.

\subsubsection{Basic Forex Indicators}
We consider several basic forex indicators like Exponential Moving Average (EMA), Moving Average Convergence Divergence (MACD), etc. prominently used by forex traders. The following crossover-based trading strategies are considered:
i. \textbf{EMA($x$)$\ \&\ $EMA($y$)}: cross-over between the EMA of past $x$ and $y$ days exchange rates, referred to as EMA($x$) and EMA($y$), respectively. In case EMA($x$) $<$ EMA($y$) at time $t$, we sell the FC revenue $R$ received till time $t$, else hold.
ii. \textbf{exchange rate$\ \&\ $EMA($x$)}
Sell at $t$ if $X_t <$ EMA($x$), else hold. 
iii. \textbf{MACD$\ \&\ $signal}
cross-over between MACD and signal where MACD is defined as the difference of EMA(12) and EMA(26), and signal is the 9-day EMA over MACD.
Accordingly, sell if MACD $<$ signal at time $t$, else hold.
iv. \textbf{MACD$\ \&\ $signal with MACD$>$0}: Sell when MACD $<$ signal as well as MACD $> 0$ at time $t$ to sell the FC revenue $R$ received till time $t$, else continue to hold.

\subsubsection{Reinforcement Learning (RL)}
We consider DQN (double deep Q-network), a deep Q-learning-based approach \cite{van2016deep} where state and action are same as defined for other approaches. 
We consider the following three variants based on definition of reward:
i. \textbf{Vanilla DQN}
with reward $r_t = a_{t}X_t$. ii. \textbf{DQN with Ranking-based Rewards}: 
Consider $\mathrm{rank}(X_t)$ to be the reverse rank (largest number gets largest value) of $X_t$ among all the exchange rates observed in an episode. Then, the reward $r_t = a_{t}\mathrm{rank}(X_t)$. iii.
\textbf{DQN with Binary Rewards}
reward $r_t=\mathds{1}_{\{a_t=a_{t^{*}}\}}$, where $a^*_t$ is the action of the Oracle policy. Here, $t$ denotes current time.

\subsubsection{Imitation Learning (IL)}
Since the oracle policy is easy to compute on a sample trajectory, we learn to imitate  it by mapping a state to optimal action using binary classification loss where the target action/class at time $t$ is $a^*_t$. In practice, the distribution of targets $a^*$ is highly skewed with lots of 0s (hold) and very few 1s (sell). Hence  we consider two variants to handle skew: i. downsample majority class, ii. use focal loss \cite{lin2017focal}. 

\subsection{Neural Network Training and Evaluation Metric} 
We use a 3-layered feedforward neural network for all learning-based approaches, where $1^{st}$ and $2^{nd}$ layers have $256$ and $128$ ReLUs, respectively. For proposed top-$K$ + AT, output units are same as $K$, for DQN-based Q-network output units are 2 (corresponding to sell/hold), and for MC formulations, network has 1 output corresponding to hold (future payoff).
We use a  learning rate of $0.003$ with Adam optimizer \cite{kingma2014adam}, batch size of $128$, and episode length $T$ of $58$. The feature vector $\mathbf{f}_{t-1}$ in I1 as well as state $s$ at time $t$ in case of the MC approaches consists of  past $n$ days FX rates, where $n \in \{5,10,20\}$. The value of $K$ for top-$K$ FX forecasts is selected from $\{1,2,3,4,5\}$. 
We consider normalized FX rates as input to the neural network, where all FX rates in an episode are divided by $X_1$. 
All the hyper-parameters across all the approaches are tuned using grid search over the validation set. For an evaluation metric, we use  average cumulative reward (ACR) defined to be $\frac{1}{E} \sum_{e \in E} \sum_{t=1}^{T} a^e_{t}X^e_{t}\Delta^e_{t}$, where $X^e_{t}$,  $a^e_{t}\in\{0,1\}$ and $\Delta^e_{t}$ are the exchange rate, action taken by trading agent, and available FC at time $t$ for an episode $e\in E$, respectively.

\subsection{Results and Observations}

\begin{table}
    \centering
	\footnotesize
	\caption{Performance comparison of proposed top-$K$ + AT (ours) with baseline Sell-AT at varying episode lengths, pointing out the robustness of the proposed approach. \label{tab:performanceLen}}
	\scalebox{0.6}{
	\centering 
		\begin{tabular}{l|c|c|c|c|c|c|c|c|c}
		\hline
		&\multicolumn{3}{|c|}{\textbf{Episode Length = 58 }} & \multicolumn{3}{|c|}{\textbf{Episode Length = 40}} & \multicolumn{3}{|c}{\textbf{Episode Length = 20}} \\ 
		 \hline
		FX-Pairs & Sell-AT & Ours & Oracle & Sell-AT & Ours & Oracle &Sell-AT& Ours & Oracle \\
		\hline
		\hline
		EURGBP  &  0.082 & \textbf{0.318}  & \textit{0.847}  &  0.062 &  \textbf{0.224} &  \textit{0.526} &  0.004 & \textbf{0.019}  & \textit{0.160}  \\
		EURUSD  & 0.643  & \textbf{0.645}  & \textit{1.194}  & 0.327  & \textbf{0.339}  & \textit{0.653}  & 0.061  & \textbf{0.098}  & \textit{0.192}  \\
		GBPUSD  & 0.870  & \textbf{1.105}  &  \textit{1.720} & 0.382  & \textbf{0.515}  & \textit{0.881}  & 0.113  & \textbf{0.137}  & \textit{0.263}  \\
		YENINR  & 0.461  &\textbf{ 0.475}  & \textit{1.245}  & 0.281  &  \textbf{0.363} & \textit{0.744}  & 0.018  & \textbf{0.050}  & \textit{0.205}  \\
		GBPINR  & 1.179  & \textbf{1.248}  & \textit{1.897}  &  0.599 & \textbf{0.626}  & \textit{0.993}  &  0.115 & \textbf{0.142}  & \textit{0.284}  \\
		USDINR  & 0.476  & \textbf{0.573}  & \textit{0.942}  & 0.244  & \textbf{0.328}  & \textit{0.528}  & 0.030  & \textbf{0.059}  & \textit{0.134}  \\
		EURINR  & \textbf{0.999}  & \textbf{0.999}  & \textit{1.573}  & \textbf{0.541}  & 0.514  & \textit{0.864}  & 0.075  & \textbf{0.107}  & \textit{0.234}  \\
		\hline
	\end{tabular}}
\end{table}

\begin{table}
    \centering
	\footnotesize
	\caption{
	Performance comparison of considering actual vs forecasted FX rates as additional inputs.  \label{tab:forecasts}}
	\scalebox{0.55}{
	\centering 
		\begin{tabular}{l|c|c|c|c|c|c|c}
		\hline
		Method &\multicolumn{1}{|c|}{\textbf{EURGBP }} & \multicolumn{1}{|c|}{\textbf{EURUSD}} & \multicolumn{1}{|c|}{\textbf{GBPUSD}} & \multicolumn{1}{|c}{\textbf{YENINR}} & \multicolumn{1}{|c}{\textbf{GBPINR}} & \multicolumn{1}{|c}{\textbf{USDINR}} & \multicolumn{1}{|c}{\textbf{EURINR}}\\
		\hline
		\hline
		Oracle & \textit{0.847} & \textit{1.194} & \textit{1.720} & \textit{1.245} & \textit{1.897} & \textit{0.942} & \textit{1.573} \\
		Oracle-5 & \textit{0.445} & \textit{0.821} & \textit{1.037} & \textit{0.724} & \textit{1.263} & \textit{0.647} & \textit{1.015} \\
		\hline
		DCNN-5 & -0.106 & \textbf{0.435} & 0.469 & \textbf{0.197} & \textbf{0.811} & 0.190 & 0.635 \\
		ARIMA-5 & \textbf{-0.087} & 0.367 & 0.489 & 0.194 & 0.767 & \textbf{0.255} & \textbf{0.696} \\
		CatBoost-5 & -0.149 & 0.368 & 0.579 & 0.062 & 0.753 & 0.112 & 0.504 \\
		LightGBM-5 & -0.144 & 0.363 & \textbf{0.581} & 0.025 & 0.780 & 0.101 & 0.501 \\
		\hline
		
		DP & \textbf{-0.070} & 0.277 & \textbf{0.388} & \textbf{0.147} & \textbf{0.546} & \textbf{0.190} & \textbf{0.454} \\
		+ Forecasts & -0.093 & 0.271 & 0.362 & 0.141 & 0.513 & 0.178 & 0.451 \\
        + Actuals & -0.074 & \textbf{0.283} & 0.382  & 0.146 & 0.543 & 0.181 & \textbf{0.454} \\
		
		\hline
        RL using DQN & \textbf{-0.031} & 0.366 & 0.458 & \textbf{0.293} & 0.664 & 0.208 & 0.559 \\
        + Forecasts & -0.243 & 0.271 & 0.408 & 0.150 & 0.607 & 0.210 & 0.547 \\
        + Actuals & -0.068 & \textbf{0.463} & \textbf{0.767} & 0.172 & \textbf{1.113} & \textbf{0.329} & \textbf{0.711} \\
        
        \hline
        Imitation Learn. & -0.050 & \textbf{0.509} & \textbf{0.442} & 0.191 & \textbf{1.070} & \textbf{0.443} & \textbf{0.942}  \\
        + Forecasts & -0.070  & 0.362 & 0.369 & 0.222 & 0.530 & 0.197 & 0.441 \\
        + Actuals & \textbf{0.102} & 0.494 & 0.416 & \textbf{0.276} & 0.865 & 0.373 & 0.736 \\
		
		\hline
		MC Q-Learn. & \textbf{0.112} & 0.492 & 0.552 & \textbf{0.349} & 0.883 & 0.374 & 0.729  \\
		+ Forecasts & -0.135 & 0.479 & 0.754 & 0.171 & 1.042 & 0.385 & 0.469 \\
        + Actuals & -0.088 & \textbf{0.560} & \textbf{0.866} & 0.221 & \textbf{1.080} & \textbf{0.414} & \textbf{0.847} \\
        
		\hline
		top-$K$ + AT & 0.318 & \textbf{0.645} & \textbf{1.105} & \textbf{0.475} & \textbf{1.248} & 0.573 & \textbf{0.999} \\
		+ Forecasts & -0.094 & 0.281 & 0.404 & 0.232 & 0.613 & 0.419 & 0.436 \\
		+ Actuals & \textbf{0.385} & 0.423 & 0.404 & 0.357 & 1.142 & \textbf{0.636} & 0.995 \\
		\hline
		
		\hline
		
	\end{tabular}}
\end{table}

Key observations from Table \ref{tab:performance}:
\newline
\textbf{1. Effectiveness of Sell-AT baseline}: Sell-AT is the most effective non-learning approach when compared against naive baselines and forex indicator strategies. 
As shown in Fig-\ref{fig:ranks_gains}(b), considering adaptive thresholds specific to individual episodes instead of one common threshold across all episodes improves performance significantly.
\newline
\textbf{2. Most learning-based approaches struggle to improve upon Sell-AT}: All learning methods using DP, RL, IL, and MC formulations perform significantly worse than Sell-AT, despite using AT as a post-processing step for final decision-making. This confirms the non-trivial nature of the learning problem at hand, and the erroneous payoff estimates due to various issues such as high degree of overestimation bias, non-stationarity, stochasticity, high class imbalance, etc.
\newline
\textbf{3. Top-$K$ FX Forecasts + AT improves upon Sell-AT}: Unlike all baselines considered, our proposed approach based on estimation of top-$K$ future FX rates followed by AT improves upon Sell-AT. It suggests that the estimate for future FX rate obtained using our approach carries useful information that improves decision-making vis-\`{a}-vis Sell-AT. 
\newline
Key observations from Table \ref{tab:performanceLen} and \ref{tab:forecasts}:
\newline
\textbf{1. Performance with varying episode lengths}: As shown in Table \ref{tab:performanceLen}, we observe that our approach performs consistently better than the second best approach, i.e. Sell-AT, indicating robustness to $T$.
\newline
\textbf{2. Vanilla $h$-step ahead Forecasting vs Proposed top-$K$ FX Forecasts:} From results in Table \ref{tab:forecasts}, we observe that the partial knowledge of actual FX rates of just the 5 future steps (we tried 1, 5, 10, and 20 future steps and had same conclusion) is sufficient to achieve performance close to Oracle that has access to all future FX rates in the episode. However, when considering 5-step ahead forecasts via traditional method like ARIMA \cite{contreras2003arima} or even learning-based methods like Deep-CNN \cite{selvin2017stock}, CatBoost \cite{dorogush2018catboost}, LightGBM \cite{sun2020novel} etc., the performance drops significantly. This shows that traditional forecasting approaches yield highly erroneous forecasts that are not effective for the problem at hand. On the other hand, our proposed approach which tries to estimate top-$K$ FX rates is significantly better than these methods. In fact, when we concatenate the forecasts as additional inputs to the historical FX rates in our method, it does not improve the performance. Instead, we observe degradation in performance indicating that standard forecasts do not carry useful signal for the task at hand.
\newline
\textbf{3. Actual vs Forecasted FX Rates as additional inputs}: We consider using forecasts or actual future FX rates as additional inputs to the neural network-based function approximators. As shown in Table \ref{tab:forecasts}, we observe that using actual forecasts can significantly improve the performance compared to the original models (without future FX rates as additional inputs) in some cases, and degrade in other cases. However, using forecasted FX rates leads to degradation in performance in most cases, again indicating the challenge in using erroneous forecasts. 
\newline
\textbf{4. Significance of $K$ in proposed approach}: As shown in Fig-\ref{fig:ranks_gains}(b), results are sensitive to $K$. We found $K>1$ to be critical to improved performance in all currency pairs (except GBPINR). One reason for this could be the fact that estimating multiple values ($K>1$) in a multi-task formulation leads to a regularization effect. In contrast,  $K=1$ would mean estimating only the maximum future FX rate which may be an outlier or not representative enough of the range of future values that FX rate can take.

\section{Conclusion}
We consider learning a trading agent for exchanging FC in the exchange market. We consider several standard formulations motivated by Optimal Stopping and their implementations with approximations across supervised learning, RL, DP, etc. We empirically show that all these solutions as well as  traditional baselines struggle to improve upon a simple baseline based on adaptive thresholds (AT). We highlight the (obvious) significance of future FX rates in good decision-making, and show that standard multistep-ahead forecasting models struggle to provide the desired benefits from estimates of future FX rates. Instead, we note that it is sufficient to correctly estimate the top-$K$ future FX rates, 
and propose a supervised learning approach for the same. Through extensive empirical evaluation, we show that our proposed approach is the only approach that is able to consistently improve upon or match the performance of the AT baseline.
It will be interesting to explore the applicability of the proposed method to other similar tasks requiring estimation of maximum/minimum future value of a variable such as buying/selling of stocks or pricing options.

\bibliography{aaai22}

\begin{thebibliography}{26}
\providecommand{\natexlab}[1]{#1}

\bibitem[{Austin et~al.(2004)Austin, Bates, Dempster~3, Leemans, and
  Williams}]{austin2004adaptive}
Austin, M.~P.; Bates, G.; Dempster~3, M.~A.; Leemans, V.; and Williams, S.~N.
  2004.
\newblock Adaptive systems for foreign exchange trading.
\newblock \emph{Quantitative Finance}, 4(4): 37--45.

\bibitem[{Becker, Cheridito, and Jentzen(2019)}]{becker2019deep}
Becker, S.; Cheridito, P.; and Jentzen, A. 2019.
\newblock Deep optimal stopping.
\newblock \emph{Journal of Machine Learning Research}, 20: 74.

\bibitem[{Bertsekas and Tsitsiklis(1995)}]{bertsekas1995neuro}
Bertsekas, D.~P.; and Tsitsiklis, J.~N. 1995.
\newblock Neuro-dynamic programming: an overview.
\newblock In \emph{Proceedings of 1995 34th IEEE Conference on Decision and
  Control}, volume~1, 560--564. IEEE.

\bibitem[{Choi, Koo, and Kwak(2004)}]{choi2004optimal}
Choi, K.~J.; Koo, H.~K.; and Kwak, D.~Y. 2004.
\newblock Optimal stopping of active portfolio management.
\newblock \emph{Annals of Economics and Finance}, 5: 119--152.

\bibitem[{Contreras et~al.(2003)Contreras, Espinola, Nogales, and
  Conejo}]{contreras2003arima}
Contreras, J.; Espinola, R.; Nogales, F.~J.; and Conejo, A.~J. 2003.
\newblock ARIMA models to predict next-day electricity prices.
\newblock \emph{IEEE Transactions on Power Systems}, 18(3): 1014--1020.

\bibitem[{Dempster and Leemans(2006)}]{dempster2006automated}
Dempster, M.~A.; and Leemans, V. 2006.
\newblock An automated FX trading system using adaptive reinforcement learning.
\newblock \emph{Expert Systems with Applications}, 30(3): 543--552.

\bibitem[{Dorogush, Ershov, and Gulin(2018)}]{dorogush2018catboost}
Dorogush, A.~V.; Ershov, V.; and Gulin, A. 2018.
\newblock CatBoost: gradient boosting with categorical features support.
\newblock \emph{arXiv preprint arXiv:1810.11363}.

\bibitem[{Duan and Kashima(2021)}]{duan2021learning}
Duan, J.; and Kashima, H. 2021.
\newblock Learning to rank for multi-step ahead time-series forecasting.
\newblock \emph{IEEE Access}, 9: 49372--49386.

\bibitem[{Dunis and Williams(2002)}]{dunis2002modelling}
Dunis, C.; and Williams, M. 2002.
\newblock Modelling and trading the EUR/USD exchange rate: Do neural network
  models perform better?
\newblock \emph{Derivatives Use, Trading and Regulation}, 8(3): 211--239.

\bibitem[{Dunis, Laws, and Sermpinis(2011)}]{dunis2011higher}
Dunis, C.~L.; Laws, J.; and Sermpinis, G. 2011.
\newblock Higher order and recurrent neural architectures for trading the
  EUR/USD exchange rate.
\newblock \emph{Quantitative Finance}, 11(4): 615--629.

\bibitem[{Fathan and Delage(2021)}]{fathan2021deep}
Fathan, A.; and Delage, E. 2021.
\newblock Deep reinforcement learning for optimal stopping with application in
  financial engineering.
\newblock \emph{arXiv preprint arXiv:2105.08877}.

\bibitem[{Fujimoto et~al.(2019)Fujimoto, Conti, Ghavamzadeh, and
  Pineau}]{fujimoto2019benchmarking}
Fujimoto, S.; Conti, E.; Ghavamzadeh, M.; and Pineau, J. 2019.
\newblock Benchmarking batch deep reinforcement learning algorithms.
\newblock \emph{arXiv preprint arXiv:1910.01708}.

\bibitem[{Fujimoto, Meger, and Precup(2019)}]{fujimoto2019off}
Fujimoto, S.; Meger, D.; and Precup, D. 2019.
\newblock Off-policy deep reinforcement learning without exploration.
\newblock In \emph{International Conference on Machine Learning}, 2052--2062.
  PMLR.

\bibitem[{Griffeath and Snell(1974)}]{griffeath1974optimal}
Griffeath, D.; and Snell, J.~L. 1974.
\newblock Optimal stopping in the stock market.
\newblock \emph{The Annals of Probability}, 1--13.

\bibitem[{Herrera et~al.(2021)Herrera, Krach, Ruyssen, and
  Teichmann}]{herrera2021optimal}
Herrera, C.; Krach, F.; Ruyssen, P.; and Teichmann, J. 2021.
\newblock Optimal stopping via randomized neural networks.
\newblock \emph{arXiv preprint arXiv:2104.13669}.

\bibitem[{Kingma and Ba(2014)}]{kingma2014adam}
Kingma, D.~P.; and Ba, J. 2014.
\newblock Adam: A method for stochastic optimization.
\newblock \emph{arXiv preprint arXiv:1412.6980}.

\bibitem[{Kohler, Krzy{\.z}ak, and Todorovic(2010)}]{kohler2010pricing}
Kohler, M.; Krzy{\.z}ak, A.; and Todorovic, N. 2010.
\newblock Pricing of high-dimensional {A}merican options by neural networks.
\newblock \emph{Mathematical Finance: An International Journal of Mathematics,
  Statistics and Financial Economics}, 20(3): 383--410.

\bibitem[{Lin et~al.(2017)Lin, Goyal, Girshick, He, and
  Doll{\'a}r}]{lin2017focal}
Lin, T.-Y.; Goyal, P.; Girshick, R.; He, K.; and Doll{\'a}r, P. 2017.
\newblock Focal loss for dense object detection.
\newblock In \emph{Proceedings of the IEEE International Conference on Computer
  Vision}, 2980--2988.

\bibitem[{Longstaff and Schwartz(2001)}]{longstaff2001valuing}
Longstaff, F.~A.; and Schwartz, E.~S. 2001.
\newblock Valuing American options by simulation: a simple least-squares
  approach.
\newblock \emph{The Review of Financial Studies}, 14(1): 113--147.

\bibitem[{Meese and Rogoff(1983)}]{meese1983empirical}
Meese, R.~A.; and Rogoff, K. 1983.
\newblock Empirical exchange rate models of the seventies: Do they fit out of
  sample?
\newblock \emph{Journal of International Economics}, 14(1-2): 3--24.

\bibitem[{Pentaliotis(2020)}]{pentaliotis2020investigating}
Pentaliotis, A. 2020.
\newblock \emph{Investigating Overestimation Bias in Reinforcement Learning}.
\newblock Ph.D. thesis.

\bibitem[{Selvin et~al.(2017)Selvin, Vinayakumar, Gopalakrishnan, Menon, and
  Soman}]{selvin2017stock}
Selvin, S.; Vinayakumar, R.; Gopalakrishnan, E.; Menon, V.~K.; and Soman, K.
  2017.
\newblock Stock price prediction using LSTM, RNN and CNN-sliding window model.
\newblock In \emph{2017 International Conference on Advances in Computing,
  Communications and Informatics (ICACCI)}, 1643--1647. IEEE.

\bibitem[{Shah and Shroff(2021)}]{shah2021forecasting}
Shah, V.; and Shroff, G. 2021.
\newblock Forecasting market prices using DL with data augmentation and
  meta-learning: {ARIMA} still wins!
\newblock \emph{arXiv preprint arXiv:2110.10233}.

\bibitem[{Sun, Liu, and Sima(2020)}]{sun2020novel}
Sun, X.; Liu, M.; and Sima, Z. 2020.
\newblock A novel cryptocurrency price trend forecasting model based on
  LightGBM.
\newblock \emph{Finance Research Letters}, 32: 101084.

\bibitem[{Van~Hasselt, Guez, and Silver(2016)}]{van2016deep}
Van~Hasselt, H.; Guez, A.; and Silver, D. 2016.
\newblock Deep reinforcement learning with double $Q$-learning.
\newblock In \emph{Proceedings of the AAAI Conference on Artificial
  Intelligence}, volume~30.

\bibitem[{Yu and Bertsekas(2007)}]{yu2007q}
Yu, H.; and Bertsekas, D.~P. 2007.
\newblock $Q$-learning algorithms for optimal stopping based on least squares.
\newblock In \emph{2007 European Control Conference (ECC)}, 2368--2375. IEEE.

\end{thebibliography}

\end{document}